\documentclass[journal,transmag]{ieeeaccess}

\ifCLASSINFOpdf

\else

\fi

\usepackage{times}
\usepackage{epsfig}
\usepackage{amsmath}
\usepackage{amssymb}

\usepackage[]{algorithm2e}

\usepackage[utf8]{inputenc}
\usepackage[english]{babel}
\usepackage{comment}
\usepackage{amsthm}
\usepackage{color}

\usepackage{array}

\usepackage{verbatim}
\usepackage[font=footnotesize]{caption}
\usepackage{subcaption}
\usepackage[switch]{lineno}
 
\definecolor{Pink}{rgb}{0.858, 0.188, 0.478}

\usepackage{cite}
\usepackage{amsmath,amssymb,amsfonts}
\usepackage{algorithmic}
\usepackage{graphicx}
\usepackage{textcomp}

\begin{document}
\history{Date of publication xxxx 00, 0000, date of current version xxxx 00, 0000.}
\doi{}

\title{Asymmetric Loss Functions and Deep Densely Connected Networks for Highly Imbalanced Medical Image Segmentation: Application to Multiple Sclerosis Lesion Detection}

\author{\uppercase{Seyed Raein Hashemi}\authorrefmark{1,2},
\uppercase{Seyed Sadegh Mohseni Salehi}\authorrefmark{1,3}, \IEEEmembership{Student Member, IEEE},
\uppercase{Deniz Erdogmus}\authorrefmark{3}, \IEEEmembership{Senior Member, IEEE},
\uppercase{Sanjay P. Prabhu}\authorrefmark{1},
\uppercase{Simon  K. Warfield}\authorrefmark{1}, \IEEEmembership{Senior Member, IEEE}, and \uppercase{Ali Gholipour}\authorrefmark{1}, \IEEEmembership{Senior Member, IEEE}}
\address[1]{Computational Radiology Laboratory, Boston Children's Hospital, and Harvard Medical School, Boston MA 02115}
\address[2]{Computer and Information Science Department, Northeastern University, Boston, MA, 02115}
\address[3]{Electrical and Computer Engineering Department, Northeastern University, Boston, MA, 02115}

\tfootnote{This study was supported in part by the National Institutes of Health grants R01 NS079788 and R01 EB018988; and in part by a Technological Innovations in Neuroscience Award from the McKnight Foundation. \\ The trained model is available as a Docker image and can be pulled with this command: docker pull \textbf{raeinhashemi/msseg2016:v1}}

\corresp{Corresponding author: Seyed Raein Hashemi (e-mail: hashemi.s@husky.neu.edu).}

\begin{abstract}

Fully convolutional deep neural networks have been asserted to be fast and precise frameworks with great potential in image segmentation. One of the major challenges in training such networks raises when data is unbalanced, which is common in many medical imaging applications such as lesion segmentation where lesion class voxels are often much lower in numbers than non-lesion voxels. A trained network with unbalanced data may make predictions with high precision and low recall, being severely biased towards the non-lesion class which is particularly undesired in most medical applications where false negatives are actually more important than false positives. Various methods have been proposed to address this problem including two step training, sample re-weighting, balanced sampling, and more recently similarity loss functions, and focal loss. In this work we trained fully convolutional deep neural networks using an asymmetric similarity loss function to mitigate the issue of data imbalance and achieve much better trade-off between precision and recall. To this end, we developed a 3D fully convolutional densely connected network (FC-DenseNet) with large overlapping image patches as input and an asymmetric similarity loss layer based on Tversky index (using $F_\beta$ scores). We used large overlapping image patches as inputs for intrinsic and extrinsic data augmentation, a patch selection algorithm, and a patch prediction fusion strategy using B-spline weighted soft voting to account for the uncertainty of prediction in patch borders. We applied this method to multiple sclerosis (MS) lesion segmentation based on two different datasets of MSSEG 2016 and ISBI longitudinal MS lesion segmentation challenge, where we achieved average Dice similarity coefficients of 69.9\% and 65.74\%, respectively, achieving top performance in both challenges. We compared the performance of our network trained with $F_\beta$ loss, focal loss, and generalized Dice loss (GDL) functions. Through September 2018 our network trained with focal loss ranked first according to the ISBI challenge overall score and resulted in the lowest reported lesion false positive rate among all submitted methods. Our network trained with the asymmetric similarity loss led to the lowest surface distance and the best lesion true positive rate that is arguably the most important performance metric in a clinical decision support system for lesion detection. The asymmetric similarity loss function based on $F_\beta$ scores allows training networks that make a better balance between precision and recall in highly unbalanced image segmentation. We achieved superior performance in MS lesion segmentation using a patch-wise 3D FC-DenseNet with a patch prediction fusion strategy, trained with asymmetric similarity loss functions.

\end{abstract}

\begin{keywords}
Asymmetric loss function, Tversky index, $F_\beta$ scores, Focal loss, Convolutional neural network, FC-DenseNet, Patch prediction fusion, Multiple Sclerosis, Lesion segmentation, Deep learning.
\end{keywords}

\maketitle

\IEEEdisplaynontitleabstractindextext

\section{Introduction}

\IEEEPARstart{C}{onvolutional} neural networks have shown promising results in a wide range of applications including image segmentation. Recent medical image processing literature shows significant progress towards automatic segmentation of brain lesions \cite{brosch2016deep,kamnitsas2017efficient}, tumors \cite{havaei2017brain,pereira2016brain,wachinger2017deepnat}, and neuroanatomy \cite{moeskops2016automatic,zhang2015deep,chen2017voxresnet} using 2D networks \cite{havaei2017brain,moeskops2016automatic,salehi2017auto}, and more recently using 3D network architectures \cite{chen2017voxresnet,kamnitsas2017efficient}. Fully convolutional networks (FCNs) with skip connections, in particular, have shown great performance  \cite{salehi2017auto,cciccek20163d,milletari2016v}. A comprehensive review of the rapidly growing literature on the broad subject of medical image segmentation requires stand-alone review papers and is beyond the scope of this article; however, after this brief introduction, we review the most relevant literature that motivated this work or is directly connected or comparable to this work in Section~\ref{sec_relatedWork}.

In this work we focus on semantic segmentation of unbalanced imaging data using deep learning, where we consider automatic brain lesion segmentation in Multiple Sclerosis (MS) as a benchmark application. MS is the most common disabling neurologic autoimmune disease resulting from recurrent attacks of inflammation in the central nervous system \cite{Steinman1996ms, Rolak2003ms}. Across the extensive literature for automated MS lesion segmentation, there are methods that try to alleviate the data imbalance issue by equal selection of training samples from each class \cite{havaei2017brain, lai2015deep}, whereas others propose using more persistent loss functions \cite{ brosch2016deep, milletari2016v, ronneberger2015u, Sudre2017gdice}.

To deal with significantly unbalanced imaging data we present \textbf{two main contributions} in this work. \textbf{First}, we propose an asymmetric loss function based on the Tversky index ($F_\beta$ scores) and compare its generality and performance to the Dice similarity loss function recently proposed for medical image segmentation using FCNs~\cite{milletari2016v}, the generalized Dice loss (GDL) function~\cite{Sudre2017gdice}, and the focal loss~\cite{lin2018focal} all proposed to deal with unbalanced data. \textbf{Second}, we proposed using large patches (as opposed to the whole image as input) that lead to relatively higher ratio of lesion versus non-lesion samples. Overlapping patches provide intrinsic data augmentation, make a better balance in data for training, and make the network adaptable for any size inputs with efficient memory usage. Despite their advantages, patches have limited effective receptive fields, therefore we propose a patch prediction fusion strategy to take into account the prediction uncertainty in patch borders. In what follows, we review the state-of-the-art in MS lesion segmentation and the related work that motivated this study. Then we show two network architectures trained with asymmetric loss functions that generate accurate lesion segmentation in ongoing MS lesion challenges and compared to the literature according to several performance metrics.

\section{Related Work}
\label{sec_relatedWork}

Many novel and genuine algorithms and models have been continuously developed and improved over the past years for MS lesion segmentation. As the number of these methods grew, so did the desire for higher precision and accuracy and more general solutions. In spite of the advances achieved by fully automated segmentation algorithms, lesion segmentation remains an active and important area of research.

The state-of-the-art MS lesion segmentation methods mostly use aggregations of skull stripping, bias correction, image registration, atlases, intensity feature information, data augmentation, and image priors or masks in training~\cite{carass2017longitudinal}. Classic supervised methods for MS lesion segmentation involved decision random forests \cite{geremia2011, jesson2015}, non-local means \cite{guizard2015rot, fartaria2016lesion}, and combined inference from patient and healthy populations~\cite{fernandez2015lesion}. More recently, deep learning methods for neural networks have shown superior performance. Among these, a recurrent
neural network with DropConnect~\cite{andermatt2017lesion}  and a cascaded convolutional neural network (CNN)~\cite{valverde2017improving} based on a cascade of two 3D patch-wise CNNs achieved remarkable results in MS lesion challenges conducted at ISBI and MICCAI conferences. In other works, two CNNs were trained sequentially in \cite{christ2017liver} where the output of the first CNN was used to select the input features of the second CNN; and a deep convolutional encoder network was proposed in \cite{brosch2016deep} where the network was pre-trained on input images using a stack of convolutional restricted boltzmann machines, and the pre-trained weights were used to initialize a convolutional two-path encoder network for fine-tuning.

In terms of deep network architectures used broadly in medical image segmentation, CNNs with independent input channel convolutions for multiple image modalities achieved state-of-the-art performance on the BRATS tumor segmentation challenge in 2016~\cite{havaei2016brain}. Segmentation techniques for 3D medical images evolved from 2D and 2.5D patchwise techniques utilizing multiple window sizes and convolutional pathways~\cite{moeskops2016automatic,kleesiek2016deep} to end-to-end segmentation using FCNs with skip connections based on U-Net~\cite{ronneberger2015u}. Autocontext networks (AutoNet)~\cite{salehi2017auto} based on a 9-pathway patchwise method as well as a U-Net style architecture showed improved brain segmentation results. A 3D extension of the U-Net, called V-Net, was suggested for medical image segmentation in~\cite{milletari2016v}. Similarly, a 3D version of the more recent densely connected architecture (DenseNet)~\cite{huang2017dense}, called DenseSeg~\cite{bui2017dense3d}, was adopted to achieve improved performance in the 2017 isointense infant brain MRI tissue segmentation challenge. A two-path fully convolutional version of the DenseNet called FC-DenseNet~\cite{jegou2017tiramisu} achieved state-of-the-art urban scene segmentation performance, and a HyperDense-Net~\cite{dolz2018hyper} architecture generated competitive segmentation results on the ongoing infant brain MRI segmentation challenge. Another recent technique~\cite{li2017hdense} merged the two popular architectures of the U-net and DenseNet while forming a hybrid structure for liver and tumor segmentation.

In addition to network architectures that evolved and used dense skip connections to improve image segmentation performance, researchers investigated the use of different loss functions. V-Net~\cite{milletari2016v} showed improved performance in unbalanced 3D medical image segmentation through training by a loss function based on the Dice similarity coefficient (DSC) compared to the widely-used cross entropy loss function.
In a recent study on lesion segmentation the generalized Dice similarity metric was suggested as a loss function (GDL)\cite{Sudre2017gdice} that outperformed weighted cross-entropy, DSC, and sensitivity-specificity loss functions. In GDL weights are assigned to different segmentation labels based on their quantity and volume in training data. Another interesting recent approach called focal loss~\cite{lin2018focal} extended the cross entropy loss function to address the class imbalance issue. It was originally used for dense object detection, however the mathematical reasoning and robustness of the function shows promise for segmentation applications as well. 

Following our preliminary work~\cite{salehi2017tversky} presented at the MICCAI 2017 Machine Learning in Medical Imaging (MLMI) workshop, where we proposed the use of the Tversky index~\cite{tversky1977features} as a loss function to train U-Net, in this paper we propose and compare the use of asymmetric similarity loss functions based on the $F_\beta$ scores (as a special case of the Tversky index, discussed in Section~\ref{sec_lossfunctions}), as well as the GDL and focal loss to train deep fully convolutional neural networks based on two network architectures: the U-net~\cite{ronneberger2015u} due to its fast speed attribute~\cite{salehi2018real} and FC-DenseNet because of its deep and powerful infrastructure \cite{huang2017dense}, both in a 3D manner. 
Through 1) training with the asymmetric similarity loss functions and 2) a 3D patch-wise approach with the FC-DenseNet method, with a patch prediction fusion strategy, all illustrated in detail as our major contributions in this paper, we achieved the best reported results in the longitudinal MS lesion segmentation challenge through September 2018 (\textcolor{blue}{https://smart-stats-tools.org/lesion-challenge}).

Within our approach, we investigated the effects of asymmetry in the similarity loss function on whole-size as well as patch-size images with two different deep networks. In addition, we incorporated a soft weighted voting method, calculating weighted average of probabilities predicted by many augmented overlapping patches in an image. Our results show that this significantly improved lesion segmentation accuracy. Based on our experimental results, we recommend the use of a 3D patch-wise fully convolutional densely connected network with large overlapping image patches and a patch prediction fusion method described here, and precision-recall balancing properties of asymmetric loss functions as a way to approach both balanced and unbalanced data in medical image segmentation where precision and recall may not have equal importance. Therefore, following our discussion on the importance of loss functions, we also present a critical discussion on the relative importance of different evaluation metrics for the applications of automatic medical image segmentation in the Discussion section. Materials and methods are presented in Section~\ref{sec_methods}, followed by results in Section~\ref{sec_results}, discussion in Section~\ref{sec_discussion}, and a conclusion in Section~\ref{sec_conclusion}.

\section{Materials and Methods}
\label{sec_methods}
\subsection{Network Architecture}
\label{sec_networks}
We trained two fully convolutional neural networks with two different network architectures: 1) a 3D fully convolutional network~\cite{long2015fully,shelhamer2017fully} based on the U-net architecture \cite{ronneberger2015u}, and 2) a 3D densely connected network \cite{huang2017dense} based on the Dense-Net architecture \cite{bui2017dense3d, jegou2017tiramisu}. To this end, we trained our 3D patch-wise FC-Dense-Net using asymmetric loss layers based on the Tversky index ($F_\beta$ scores), focal loss, and generalized Dice similarity coefficient (GDL), and compared it against a 3D-Unet with asymmetric loss layer. The details of the network architectures are described next and we follow with the loss function formulation, and our proposed 3D patch prediction fusion method for the patch-wise network.

\begin{table}
\small
\centering
\caption{Architecture details of 3D FC-DenseNet. Each Dense Block consists of four groups of $1\times1\times1$ convolutions as bottlenecks plus $3\times3\times3$ convolutions where all convolutional layers are followed by batch normalization and ReLU activation layers. Transition Down blocks consist of $1\times1\times1$ convolutions and $2\times2\times2$ max pooling layers, while transition up blocks consist of $3\times3\times3$ transpose convolutions with strides of $2\times2\times2$.}
\label{table:FCDenseNet}
 \begin{tabular}{|c|} 
 \hline
  \textbf{Architecture}\\
  \hline
  \hline
  Input ($64\times64\times64$) or ($128\times128\times128$)\\
  with 4 or 5 modalities \\
  \hline
  $2\times2\times2$ Convolution + strides of $2\times2\times2$ \\
  (only for $128\times128\times128$ input patches)\\
  \hline
  $3\times3\times3$ Convolution + BN + ReLU\\
  \hline
  $3\times3\times3$ Convolution + BN + ReLU\\
  \hline
  $3\times3\times3$ Convolution + BN + ReLU\\
  \hline
  Dense Block (4 bottleneck + 4 conv layers) + Transition Down \\
  \hline
  Dense Block (4 bottleneck + 4 conv layers) + Transition Down \\
  \hline
  Dense Block (4 bottleneck + 4 conv layers) + Transition Down \\
  \hline
  Dense Block (4 bottleneck + 4 conv layers) + Transition Down \\
  \hline
  Dense Block (4 bottleneck + 4 conv layers) + Transition Down \\
  \hline
  Dense Block (4 bottleneck + 4 conv layers) \\
  \hline
   Transition Up + Dense Block (4 bottleneck + 4 conv layers) \\
   \hline
   Transition Up + Dense Block (4 bottleneck + 4 conv layers) \\
   \hline
   Transition Up + Dense Block (4 bottleneck + 4 conv layers) \\
   \hline
   Transition Up + Dense Block (4 bottleneck + 4 conv layers) \\
   \hline
   Transition Up + Dense Block (4 bottleneck + 4 conv layers) \\
   \hline
   $2\times2\times2$ Transpose Conv + strides of $2\times2\times2$ \\
  (only for $128\times128\times128$ input patches)\\
   \hline
  $1\times1\times1$ Convolution\\
  \hline
  Sigmoid\\
   \hline
\end{tabular}
\end{table}

\begin{figure*}
    \centering
    \includegraphics[width=0.95\textwidth]{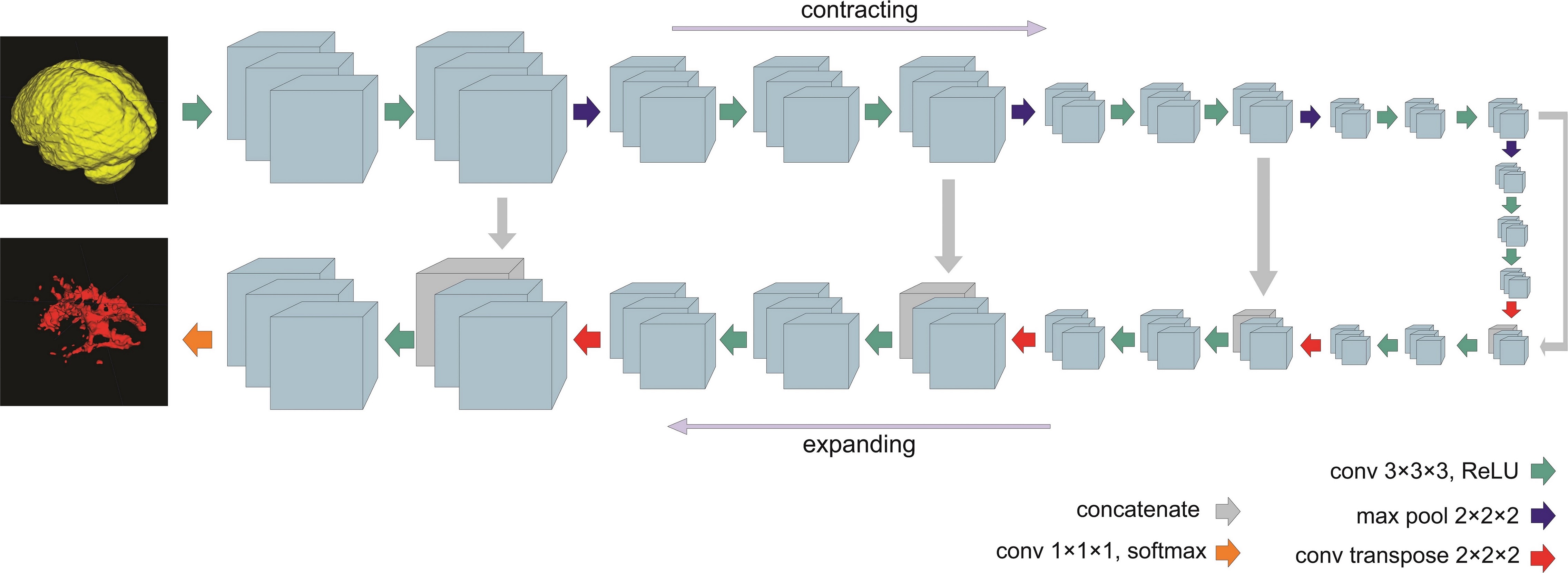}
    \caption{The 3D U-net style architecture with full-size multi-channel images as inputs and skip connections between a contracting path and an expanding path.}
    \label{fig:Net}
\end{figure*}

\begin{figure*}
    \centering
    \includegraphics[width=0.95\textwidth]{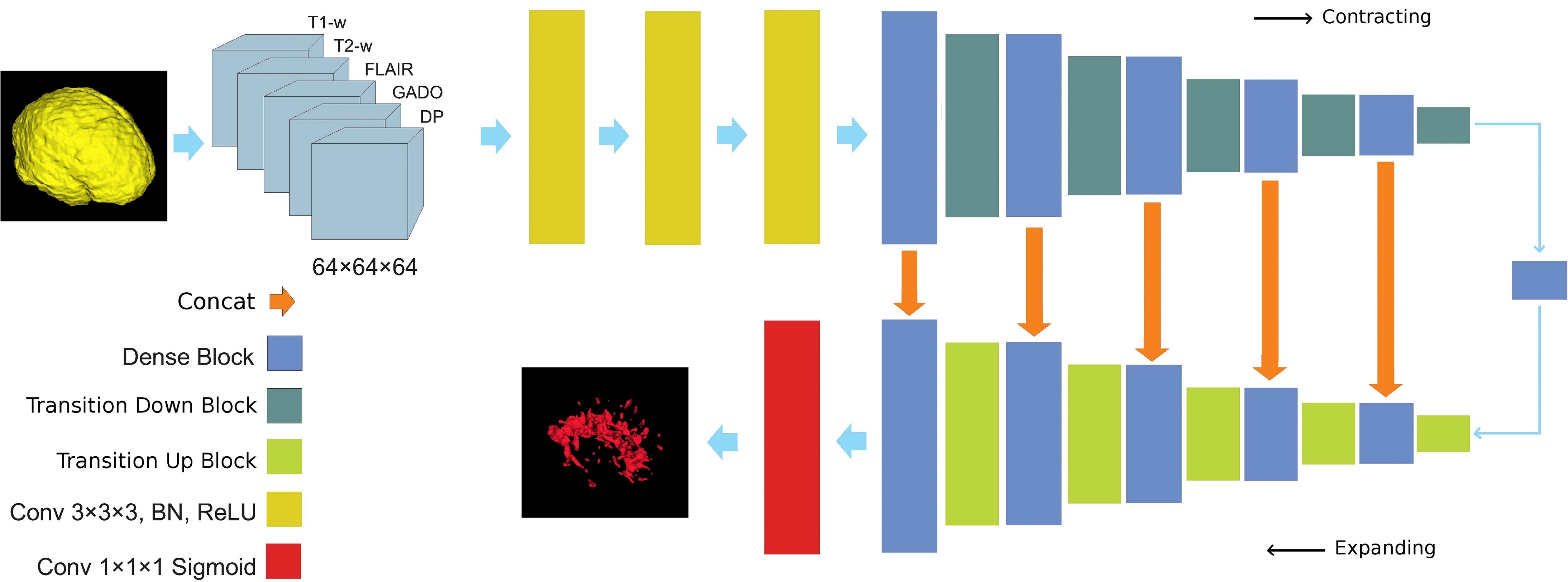}
    \caption{The 3D patch-wise FC-Dense-Net style architecture with $64\times64\times64$ five channel input patches, consisting of eleven dense blocks and four convolutional layers with bottlenecks within each block. Overlapping patches of a full size image are used as inputs to this network for training and testing.}
    \label{fig:Dense}
\end{figure*}

\subsubsection{3D U-net}
As the baseline, we train a 3D U-net with an asymmetric similarity loss layer~\cite{salehi2017tversky}. This U-net style architecture 
is shown in Figure~\ref{fig:Net}. It consists of a contracting and an expanding path (to the right and left, respectively). High-resolution features in the contracting path are concatenated with upsampled versions of global low-resolution features in the expanding path to help the network learn both local and global information. 
In the contracting path, padded $3\times3\times3$ convolutions are followed by ReLU non-linear layers. $2\times2\times2$ max pooling with stride 2 is applied after every two convolutional layers. The number of features is doubled after each downsampling by the max pooling layers. The expanding path contains $2\times2\times2$ transposed convolution layers after every two convolutional layers, and the resulting feature map is concatenated to the corresponding feature map from the contracting path. At the final layer a $1\times1\times1$ convolution with softmax activation is used to reach the feature map with depth of two, equal to the number of lesion and non-lesion classes.

\subsubsection{3D Patch-Wise FC-DenseNet}

We propose a 3D patch-wise fully convolutional Dense-Net based on 3D DenseSeg \cite{bui2017dense3d} and FC-DenseNet \cite{jegou2017tiramisu} with overlapping patches, an asymmetric similarity loss layer and a patch prediction fusion strategy. The focal loss~\cite{lin2018focal} and generalized Dice loss~\cite{Sudre2017gdice} (GDL) functions were also used to train this architecture for comparison purposes. Figure~\ref{fig:Dense} shows the schematic architecture of the 3D patch-wise FC-DenseNet, and Table~\ref{table:FCDenseNet} shows the details of this network architecture. This Dense-Net style architecture consists of three initial $3\times3\times3$ convolutional layers followed by five dense blocks in the contracting path, a dense block in the middle, and another five dense blocks in the expanding path. Growth rate of 12 is applied to each dense block in the network. Growth rate refers to the increase amount in the number of feature maps after each layer in a dense block. In each dense block there are four $3\times3\times3$ convolutional layers preceding with $1\times1\times1$ convolutional layers referred to as bottlenecks \cite{huang2017dense}, which have the purpose of reducing the number of input feature maps. Skip connections are made between all layers of each dense block. Aside from the center dense block connecting the two paths, dense blocks in the contracting path are followed by a $1\times1\times1$ convolutional layer and a max pooling layer named transition down blocks and dense blocks in the expanding path are preceded with $3\times3\times3$ transpose convolution layers of stride 2 known as transition up blocks~\cite{jegou2017tiramisu}. Dimension reduction of 0.5 is applied at transition layers to help reduce the feature map dimensionality for computational and parameter efficiency. Each of the convolutional layers is followed by batch normalization and ReLU activation layers. Dropout rate of 0.2 is only applied after $3\times3\times3$ convolutional layers within dense blocks. At the final layer a $1\times1\times1$ convolution with sigmoid output is used to reach the feature map with depth of one (lesion or non-lesion class).

Prior to proceeding to the main classifier, results of all dense blocks are upsampled using deconvolutional layers, using transpose matrices of convolutions. Afterwards, the results are concatenated and passed through the main classifier to calculate the probability map of the input patch. In the proposed architecture, fully convolutional layers are used instead of fully connected layers~\cite{sermanet2013overfeat} to achieve much faster testing time. This architecture segments large 3D image patches. Therefore, to segment any size input image, overlapping large patches (typically of size $64\times64\times64$ or $128\times128\times128$) extracted from the image are used as input to the network. These patches are augmented and their predictions are fused to provide final segmentation of a full-size input image. The loss layer, patch augmentation and patch prediction fusion, and the details of training are discussed in the sections that follow.

\subsection{Asymmetric Similarity Loss Function}
\label{sec_lossfunctions}
The output layers in our two networks consist of 1 plane. There is one plane for the MS Lesion class. Lesion voxels are labeled as 1 and non-lesion voxels are labeld as zero. We applied sigmoid on each voxel in the last layer to form the last feature map. Let $P$ and $G$ be the set of predicted and ground truth binary labels, respectively. The Dice similarity coefficient $D$ between $P$ and $G$ is defined as:
\begin{equation}
    D(P,G) = \frac{2|PG|}{|P|+|G|}
\end{equation}
Loss functions based on the Dice similarity coefficient have been proposed as alternatives to cross entropy to improve training 3D U-Net (V-net) and other network architectures~\cite{milletari2016v,Sudre2017gdice}; however $D$, as the harmonic mean of precision and recall, weighs false positives (FPs) and false negatives (FNs) equally, forming a symmetric similarity loss function. To make a better adjustment of the weights of FPs and FNs (and achieve a better balance between precision and recall) in training fully convolutional deep networks for highly unbalanced data, where detecting small number of voxels in a class is crucial, we propose an asymmetric similarity loss function based on the $F_\beta$ scores which is defined as:
\begin{equation}
    F_\beta = (1+\beta^2) \frac{precision \times recall}{\beta^2 \times precision+recall}
\label{eq:fbeta}
\end{equation}
Equation~(\ref{eq:fbeta}) can be written as:
\begin{equation}
    F(P,G;\beta) = \frac{(1+\beta^2)|PG|}{(1+\beta^2)|PG|+\beta^2 |G\setminus P|+|P\setminus G|}
\label{eq:fbeta_tp}
\end{equation}
\noindent where $|P\setminus G|$ is the relative complement of $G$ on $P$. To define the $F_\beta$ loss function we use the following formulation:
\begin{flalign*}
F_\beta &= &
\end{flalign*}
\small
\begin{equation}
\begin{split}
    \frac{(1+\beta^2) \sum_{i=1}^Np_{i}g_{i}}{(1+\beta^2) \sum_{i=1}^Np_{i}g_{i} + \beta^2 \sum_{i=1}^N(1-p_{i})g_{i} + \sum_{i=1}^Np_{i}(1-g_{i})}
    \label{fbeta_loss}
\end{split}
\end{equation}
\normalsize
\noindent where in the output of the sigmoid layer, the $p_{i}$ is the probability of voxel $i$ be a lesion and $1-p_{i}$ is the probability of voxel $i$ be a non-lesion. Additionally, the ground truth training label $g_{i}$ is 1 for a lesion voxel and 0 for a non-lesion voxel. The gradient of the $F_\beta$ in Equation~(\ref{fbeta_loss}) with respect to $P$ is defined as 
$\nabla F_\beta=[\frac{\partial F_\beta}{\partial p_{1}}, \frac{\partial F_\beta}{\partial p_{2}}, ..., \frac{\partial F_\beta}{\partial p_{N}}]$  
where each element of gradient vector can be calculated as:
\begin{flalign*}
\frac{\partial F_\beta}{\partial p_{j}} &= &
\end{flalign*}
\small
\begin{equation}
  \frac{(1+\beta^2)g_j(\beta^2 \sum_{i=1}^N(1-p_{i})g_{i} + \sum_{i=1}^Np_{i}(1-g_{i}))}{((1+\beta^2) \sum_{i=1}^Np_{i}g_{i} + \beta^2 \sum_{i=1}^N(1-p_{i})g_{i} + \sum_{i=1}^Np_{i}(1-g_{i}))^2}
\end{equation}
\normalsize
Considering this formulation we do not need to use weights to balance the training data. Also by adjusting the hyperparameter $\beta$ we can control the trade-off between precision and recall (FPs and FNs). For better interpretability to choose $\beta$ values, we rewrite Equation~(\ref{eq:fbeta_tp}) as
\begin{equation}
    F(P,G;\beta) = \frac{|PG|}{|PG|+\frac{\beta^2}{(1+\beta^2)} |G\setminus P|+\frac{1}{(1+\beta^2)}|P\setminus G|}
\label{eq:fbeta_tp2}
\end{equation}

It is notable that the $F_\beta$ index is a special case of Tversky index~\cite{tversky1977features}, where the constraint $\alpha + \beta = 1$ is preserved. The asymmetric $F_\beta$ loss function with the hyper-parameter $\beta$ generalizes the Dice similarity coefficient and the Tanimoto coefficient (also known as the Jaccard index). More specifically, in the case of $\beta = 1$ the $F_\beta$ index simplifies to be the Dice loss function ($F_1$) while $\beta = 2$ generates the $F_2$ score and $\beta = 0$ transforms the function to precision. Larger $\beta$ weighs recall higher than precision (by placing more emphasis on false negatives). We hypothesize that using higher $\beta$ in the asymmetric similarity loss function helps us shift the emphasis to decrease FNs and boost recall, therefore achieve better performance in terms of precision-recall trade-off. Appropriate values of the hyper-parameter $\beta$ can be defined based on class imbalance ratios.

\subsection{3D Patch Prediction Fusion}
\label{sec_patches}

To use our 3D patch-wise FC-DenseNet architecture to segment a full-size input image (of any size), overlapping large patches (of size $64\times64\times64$ or $128\times128\times128$) are taken from the image and fed into the network. In both training and testing, patches are augmented, fed into the network, and their predictions are fused in a procedure that is described in this section. A network with smaller input patch size uses less memory. Therefore, to fit the $128\times128\times128$ size patches into the memory we used an extra $2\times2\times2$ convolution layer with stride 2 at the very beginning of our architecture to reduce the image size.

\begin{figure}
    \centering
    \includegraphics[width=1\columnwidth]{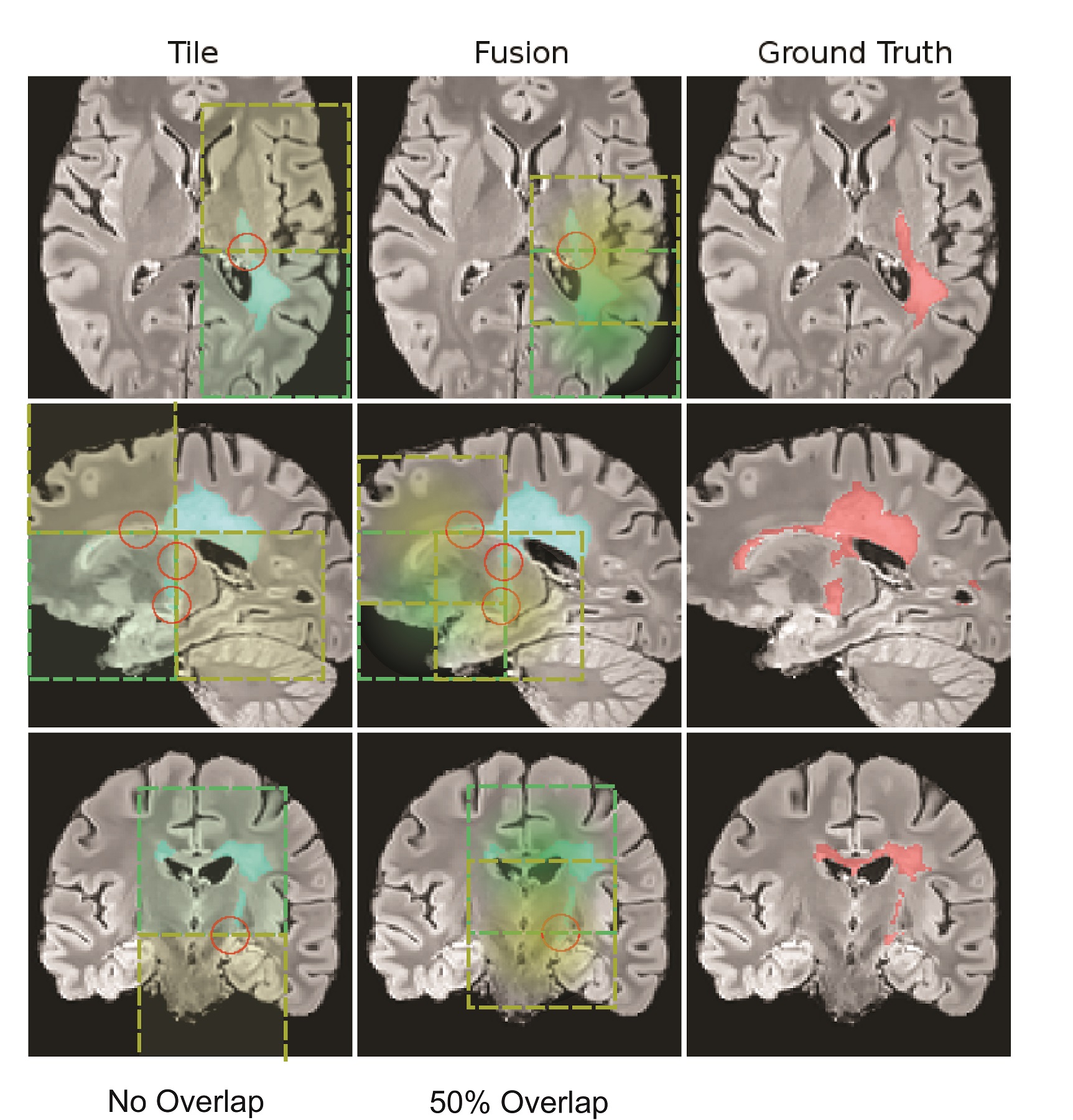}
    \caption{The 2D representation of patch selection in the 3D patch fusion method compared to the 3D patch tiling method. Each voxel is covered by 8 3D patches in the fusion approach while only predicted once in the tiling approach. The presented predictions are based on the FC-DenseNet model trained with the asymmetric similarity loss function with $\beta=1.5$. Voxels near patch borders get relatively lower accuracy predictions when a tiling approach is used, while for the fusion approach voxels near the border of one patch will be at the center of at least another patch resulting in a higher accuracy. The differences of predictions are marked with red circles as they are compared with the ground truth on the right. Please note that 3D spline function weights are displayed in 2D as transparent green and yellow highlight colors.}
    \label{fig:win}
\end{figure}

The amount of intersection area (overlap) between patches is adjustable. For example for images of size $128\times224\times256$, the prediction time using 75\% overlaps was roughly 45 minutes per 3D image. However, to keep the prediction time close to 5 minutes per image, we used 50\% overlaps (stride of 1/2 of the patch size) on patch windows. Therefore, given input image sizes of $128\times224\times256$, for example, the algorithm produces $5\times8\times9$ patches per augmentation (5 patches of $64\times64\times64$ along coronal, 8 patches of $64\times64\times64$ along sagittal and 9 patches of $64\times64\times64$ along axial planes). There are four augmentations, the original image, and the three 180 degree rotations for each plane. Consequently, our model performs 1,440 3D patch predictions of size $64\times64\times64$ per 3D image (of the above-mentioned size) leading to 32 prediction probabilities per voxel which later are used in soft voting. If we were to use 75\% overlaps, we had to predict 2,295 ($9 \times 15 \times 17$) number of 3D patches multiplied by 4 per augmentation (9,180 3D patch predictions in total), which is more than 6 times the number of patches to predict for 50\% overlaps (360 patches of $5\times8\times9$ leading to 1,440 3D patch predictions in total). The testing time for all images  of different sizes that we considered in this study was between 1 to 5 minutes with the 50\% overlap for 32 overlapping patch predictions.

The predictions from overlapping patches are fused to form the segmentation of the full-size image. In case of no overlap and no patch augmentation, each voxel on the original image has only one predicted value (compared to 32 predictions per voxel in 50\% overlapping patches), therefore predictions from patches can just be tiled to produce the original image segmentation. However, this does not lead to the best results due to the lack of augmentation in test and training and also because patch predictions are less accurate in the patch borders due to limited effective receptive field~\cite{luo2016understanding} of patches and incomplete image features in patch borders. The 2D representation of the effect is shown in Figure~\ref{fig:win} where lesions in the border of patches are not correctly segmented in the tiling method where no overlap between patches was used. In the second column, where patches with 50\% overlap were used, each voxel received multiple predictions from overlapping patches (not all 8 3D patches per voxel could be displayed in the 2D figure).

To take into account the relative uncertainty of predictions near patch borders, we use weighted soft voting to fuse patch predictions as opposed to the conventional voting (averaging) method~\cite{bernal2018quantitative}. To this end, we calculate the relative weights of soft predictions using a second-order spline function at each patch center as an efficient and compact model~\cite{wang1998scale} of Gaussian effective receptive fields~\cite{luo2016understanding}. This allows fusion of predictions from all overlapping and augmented patches while giving lower weights to predictions made at patch borders. With 50\% overlap, voxels near the borders of one patch are near the center of another patch as seen in Figure~\ref{fig:win}. In our experiments we compared different scenarios, in particular compared our proposed spline patch prediction fusion with uniform patch prediction fusion and patch tiling. It is noteworthy that all selected patches, overlaps, spline functions and soft voting methods are performed in 3D and all 2D representations are actually fulfilled in a 3D manner.

\subsection{Datasets}
We trained and evaluated our networks on data sets from the MS lesion segmentation (MSSEG) challenge of the 2016 Medical Image Computing and Computer Assisted Intervention conference~\cite{commowick2016msseg} (\textcolor{blue}{https://portal.fli-iam.irisa.fr/msseg-challenge/overview}) as well as the longitudinal MS lesion segmentation challenge of the IEEE International Symposium on Biomedical Imaging (ISBI) conference~\cite{carass2017longitudinal} (\textcolor{blue}{https://smart-stats-tools.org/lesion-challenge)}, which is an ongoing challenge with lively reported results on test data. T1-weighted magnetization prepared rapid gradient echo (MPRAGE), Fluid-Attenuated Inversion Recovery (FLAIR), Gadolinium-enhanced T1-weighted MRI, Proton Density (PD), and T2-weighted MRI scans of 15 subjects were used as five channel inputs for the MSSEG challenge, and T1-weighted MPRAGE, FLAIR, PD, and T2-weighted MRI scans of 5 subjects with a total of 21 stacks were used as four channel inputs for the ISBI challenge.
In the MSSEG dataset, every group of five subjects were in different domains: 1) Philips Ingenia 3T, 2) Siemens Aera 1.5T and 3) Siemens Verio 3T. In the ISBI dataset, all scans were acquired on a 3.0 Tesla MRI scanner.
Images of different sizes were all rigidly registered to a reference image of size $128 \times 224 \times 256$ for the MSSEG dataset. After registration, average lesion voxels per image was 15,500, with a maximum of 51,870 and a minimum of 647 voxels.

\subsection{Training}
We trained our two FCNs with asymmetric loss layers to segment MS lesions in MSSEG and ISBI datasets.  Both datasets were trained with a five-fold cross-validation strategy where five instances of training were performed on $4/5^{th}$ of each dataset and validated on the remaining $1/5^{th}$. Therefore, in each turn we trained our network on 12 subjects and validated on 3 subjects in the MSSEG dataset, while in the ISBI dataset 17 stacks were used for training and 4 for validation. Details of the training process of each network are described here.

\subsubsection{3D Unet}
The 3D U-Net was trained end-to-end. Cost minimization on 1000 epochs was performed on the MSSEG dataset using ADAM optimizer~\cite{kingma2014adam} with an initial learning rate of 0.0001 multiplied by 0.9 every 1000 steps. The training time for this network was approximately 4 hours on a workstation with Nvidia Geforce GTX1080 GPU.

\subsubsection{3D patch-wise FC-DenseNet}
Our 3D patch-wise FC-DenseNet was trained end-to-end. Cost minimization on 4000 epochs (for the MSSEG dataset) and 1000 epochs (for the ISBI dataset) was performed using ADAM optimizer~\cite{kingma2014adam} with an initial learning rate of 0.0005 multiplied by 0.95 every 500 steps with a step growth rate of 2 every 16,000 steps. For instance, the first growth happens at the 16,000th step, where the interval of 500 would be multiplied by two. The training time for this network was approximately 16 hours (MSSEG) and 4 hours (ISBI) on a workstation with Nvidia Geforce GTX1080 GPU. The input patch size was chosen $64\times64\times64$ for the MSSEG images and $128\times128\times128$ for the ISBI images in a trade-off between accuracy of extracted features (field-of-view) in each patch and limitations on the GPU memory. The selected size appeared to be both effective and practical for comparisons.

Similarity loss functions (including the Dice similarity coefficient and our proposed asymmetric similarity loss) rely on true positive (TP) counts. The networks would not be able to learn if the TP value is zero leading to a zero loss value. Therefore, only patches with a minimum of 10 lesion voxels were selected for training the patch-wise FC-DenseNet architecture. Nevertheless, equal number of patches was selected from each image. Therefore, the FCNs trained equally with the training data, although they may have had a more diverse pool on images with more number of lesion voxels. This network was trained with our proposed asymmetric similarity loss function as well as the GDL and focal loss functions. The results of these methods can be found in the following sections.

\subsection{Testing} 
In order to train and test the architectures properly, five-fold cross validation was used as the total number of subjects was very limited. For MSSEG dataset, each fold contained 3 subjects each from 3 different centers. For ISBI dataset, each fold contained 4 stacks from one subject (total of 5 subjects). In order to test each fold we trained the networks each time from the beginning using the other 4 folds containing images of 12 subjects (MSSEG) and 4 subjects with 4 stacks each (ISBI). After feeding forward the test subjects through the networks, voxels with computed probabilities of 0.5 or more were considered to belong to the lesion class and those with probabilities $<0.5$ were considered non-lesion. The results of our trained models with focal loss, GDL, and the asymmetric similarity loss functions for the test data were submitted through the ISBI challenge portal, and the results appeared on the publicly available results board (\textcolor{blue}{https://smart-stats-tools.org/lesion-challenge}). 

\section{Experiments and Results}
\label{sec_results}
We conducted experiments to evaluate the relative effectiveness of different networks, asymmetry in loss functions, and patch prediction fusion on lesion segmentation. In this section, first we describe the wide range of metrics used for evaluation, and then present the results of experiments on the two challenge datasets, where we compare our methods with the results reported in the literature, and in the challenge.

\subsection{Evaluation Metrics}
To evaluate the performance of our networks and compare them against state-of-the-art methods in MS lesion segmentation, we calculate and report several metrics including those used in the literature and the challenges. This includes the Dice Similarity Coefficient (DSC) which is the ratio of twice the amount of intersection to the total number of voxels in prediction ($P$) and ground truth ($G$), defined as:
\begin{flalign*}
DSC = \frac{2\left | P\cap G \right |}{\left | P \right |+\left | G \right |} = \frac{2TP}{2TP+FP+FN} &
\end{flalign*}
where $TP$, $FP$, and $FN$ are the true positive, false positive, and false negative rates, respectively. We also calculate and report sensitivity (recall) defined as $\frac{TP}{TP+FN}$ and specificity defined as $\frac{TN}{TN+FP}$ and the $F_2$ score as a measure that is commonly used in applications where recall is more important than precision (as compared to $F_1$ or DSC): 
\begin{flalign*}
F_2 = \frac{5TP}{5TP+4FN+FP}
\end{flalign*}

To critically evaluate the performance of lesion segmentation for the highly unbalanced (skewed) datasets, we use the Precision-Recall (PR) curve (as opposed to the receiver-operator characteristic, or ROC, curve) as well as the area under the PR curve (the APR score)~\cite{boyd2013area,davis2006relationship,fawcett2006introduction}. For such skewed datasets, the PR curves and APR scores (on test data) are preferred figures of algorithm performance.

In addition to DSC and True Positive Rate (TPR, same as sensitivity or recall), seven other metrics were used in the ISBI challenge. These included the Jaccard index defined as:
\begin{flalign*}
Jaccard = \frac{TP}{TP+FP+FN}
\end{flalign*}
the Positive Predictive Value (PPV) defined as the ratio of true positives to the sum of true and false positives:
\begin{flalign*}
PPV = \frac{TP}{TP+FP}
\end{flalign*}
the lesion-wise true positive rate (LTPR), and lesion-wise false positive rate (LFPR), which are more sensitive in measuring
the accuracy of segmentation for smaller lesions that are important to detect when performing early disease diagnosis \cite{garcia2013lesion}. LTPR is the ratio of true positives to the sum of true positives and false negatives, whereas LFPR is the ratio of false positives to the sum of false positives and true negatives, both only on lesion voxels:
\begin{flalign*}
LTPR = \frac{TP}{TP+FN} \quad, \quad 
LFPR = \frac{FP}{FP+TN}
\end{flalign*}
the Volume Difference (VD) defined as the absolute difference in volumes divided by the volume of ground truth:
\begin{flalign*}
VD = \frac{Vol(Seg) - Vol(GT)}{Vol(GT)}
\end{flalign*}
where GT and Seg denote ground truth and predicted segmentation, respectively; the average segmentation volume which is the average of all segmented lesion volumes; and the average symmetric Surface Difference ($SD$) which is the average of the distance (in millimetres) from the predicted lesions to the nearest GT lesions plus the distance from the GT lesions to the nearest predicted lesions~\cite{carass2017longitudinal}. A value of $SD=0$ would correspond to identical predicted and ground truth lesions.

An overall score is also calculated in ISBI challenge based on a combination of these metrics; however, it has been mentioned~\cite{carass2017longitudinal} that this single score does not necessarily represent the best criteria. We will discuss the criteria and performance in terms of individual metrics in the sections that follow. In particular, in the discussion section, we will discuss why some performance criteria are more important than others for applications such as disease diagnosis and prognosis based on lesion detection for clinical judgment and treatment planning. These criteria were central to our objectives that led to the proposed strategies to effectively train deep neural networks for improved lesion segmentation. 

\begin{figure*}
    \centering
    \includegraphics[width=1\textwidth]{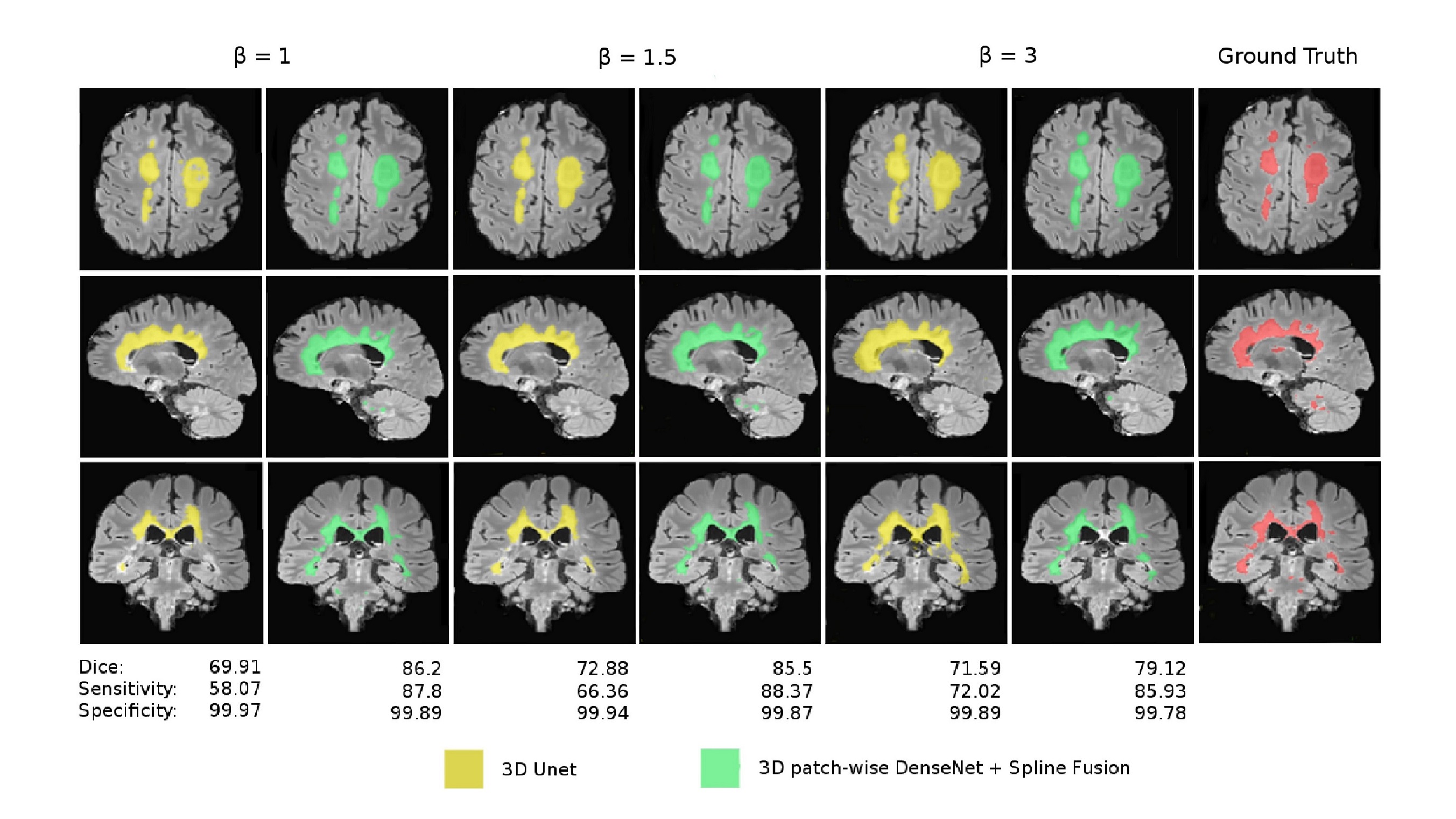}
    \caption{The effect of different weights on FP and FN imposed by the asymmetric loss function on a case with extremely high density of lesions. Axial, sagittal, and coronal sections of images have been shown and the Dice, sensitivity, and specificity values of each case are shown underneath the corresponding column. The best results were obtained at $\beta=1.5$ with our proposed 3D patch-wise FC-DenseNet with spline patch prediction fusion.}
    \label{fig:HighL}
\end{figure*}

\begin{figure*}
    \centering
    \includegraphics[width=1\textwidth]{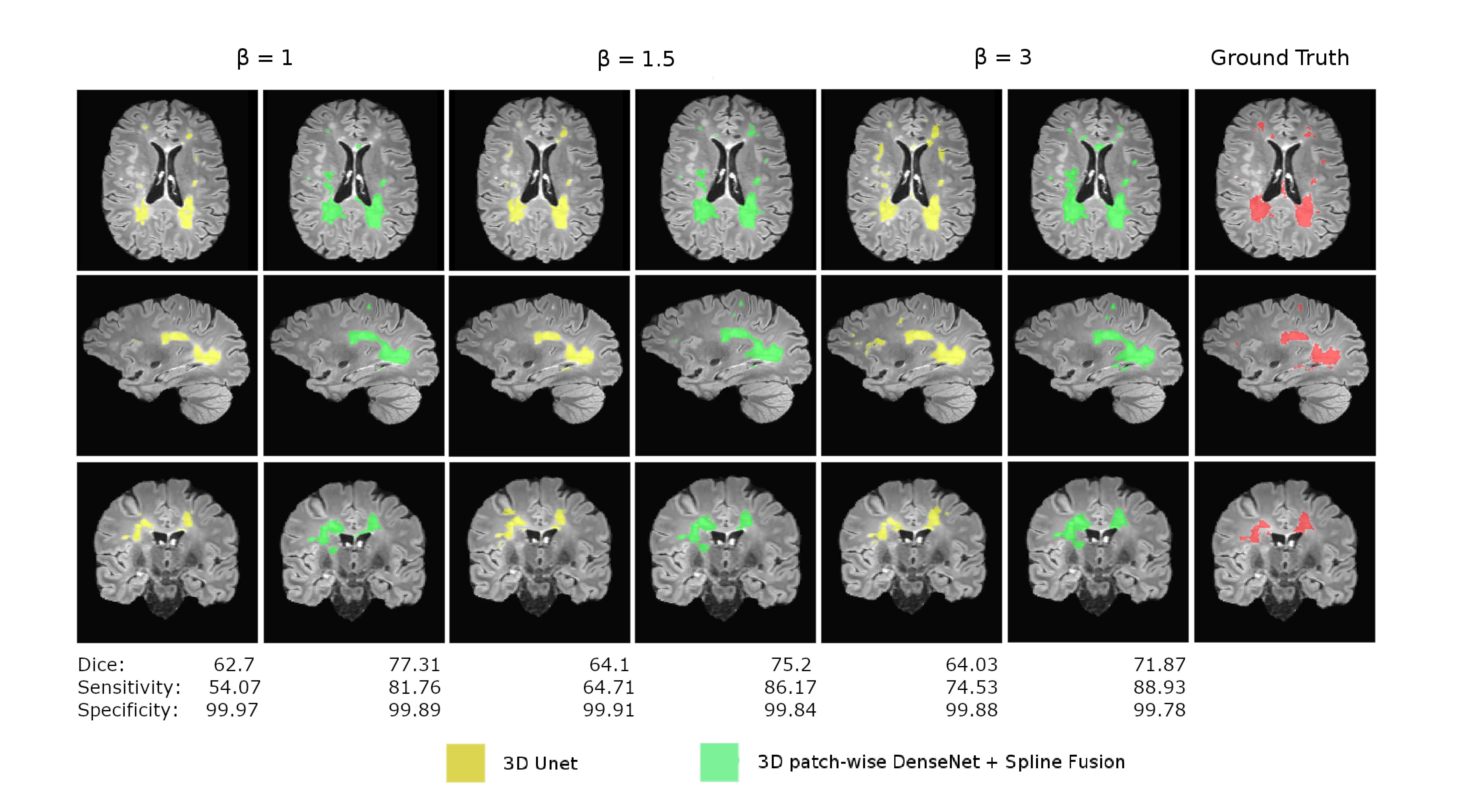}
    \caption{The effect of different weights on FP and FN imposed by the asymmetric loss function on a case with medium density of lesions. Axial, sagittal, and coronal sections of images have been shown and the Dice, sensitivity, and specificity values of each case are shown underneath the corresponding column. The best results were obtained at $\beta=1.5$ with our proposed 3D patch-wise FC-DenseNet with spline patch prediction fusion.}
    \label{fig:MediumL}
\end{figure*}

\begin{figure*}
    \centering
    \includegraphics[width=1\textwidth]{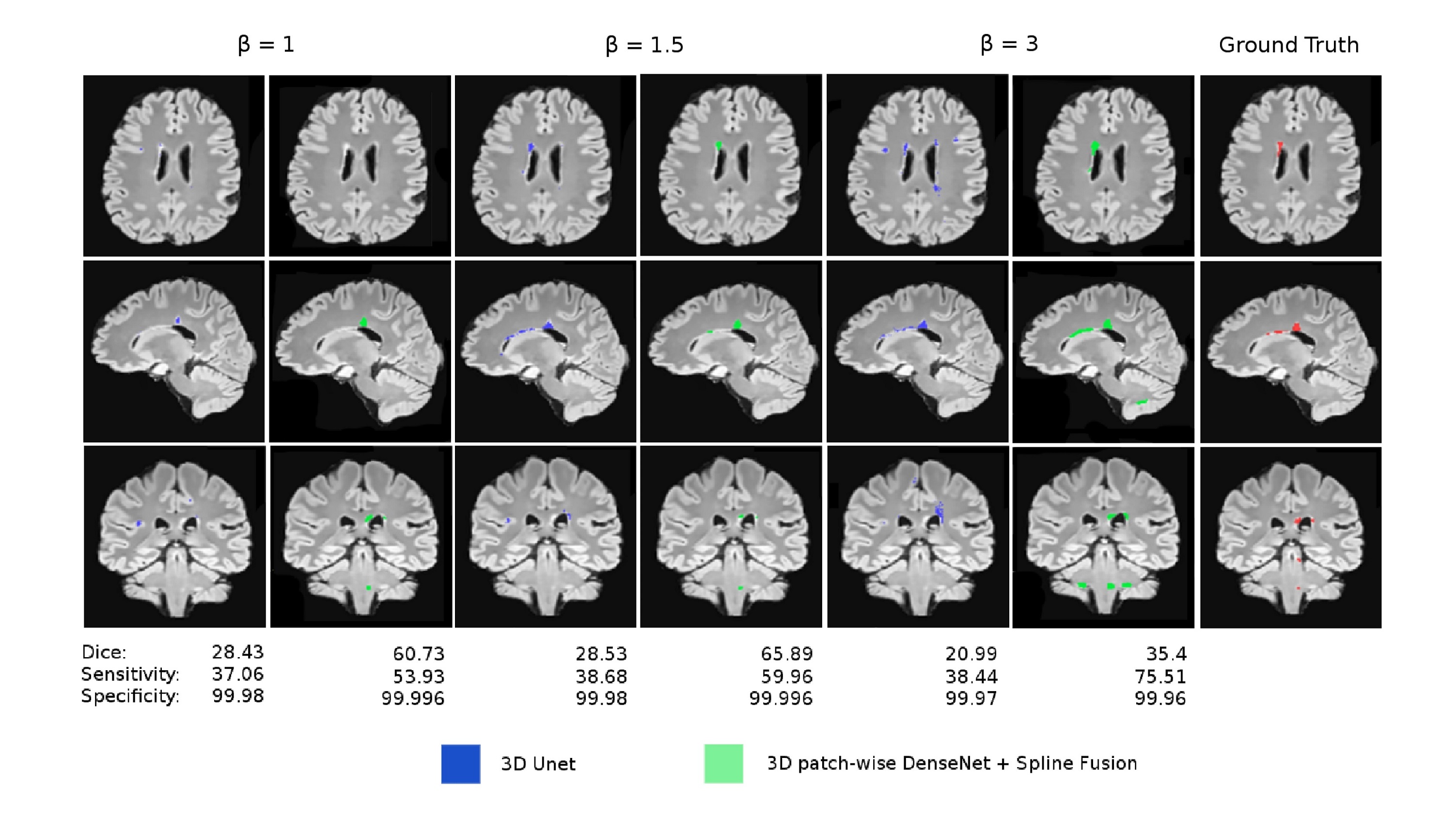}
    \caption{The effect of different weights on FP and FN imposed by the asymmetric loss function on a case with extremely low density of lesions. Axial, sagittal, and coronal sections of images have been shown and the Dice, sensitivity, and specificity values of each case are shown underneath the corresponding column. The best results were obtained at $\beta=1.5$ with our proposed 3D patch-wise FC-DenseNet with spline patch prediction fusion.}
    \label{fig:LowL}
\end{figure*}

\subsection{Results}
\subsubsection{Evaluation on the MSSEG dataset}
To evaluate the effect of the asymmetric loss function in making the trade-off between precision and recall, and compare it with the Dice loss function (which is the harmonic mean of precision and recall) in MS lesion segmentation, we trained our FCNs with different $\beta$ values on the MSSEG dataset. Note that $\beta=1$ in Equation~(\ref{eq:fbeta_tp}) corresponds to the Dice loss function. Based on equation \ref{eq:fbeta_tp2}, we chose $\beta$s so that the coefficient of $|G\setminus P|$ (false negatives) spanned over 0.5 to 0.9 with an interval of 0.1 in our tests.
The performance metrics are reported in Table~\ref{table:res}. These results show that 1) the balance between sensitivity and specificity was controlled by the parameters of the loss function; 2) according to all combined test measures (i.e. DSC, $F_2$, and APR score), the best results were obtained from the FCNs trained with $\beta=\sqrt{\frac{7}{3}}\sim1.5$, which performed better than the FCNs trained with the Dice loss function corresponding to $\beta=1$; 3) the results obtained from 3D patch-wise FC-DenseNet were much better than the results obtained from 3D U-net; and 4) our proposed spline fusion of patch predictions led to improved performance of the patch-wise FC-DenseNet with tiling and uniform patch prediction fusion. Overall, the best results were obtained with the 3D patch-wise FC-DenseNet with asymmetric loss at $\beta=1.5$, and spline-weighted soft voting for patch prediction fusion. It is also noteworthy that the gain in performance due to individual components of our algorithm, in particular the asymmetric similarity loss, with the most complex, best performing model (3D patch-wise FC-DenseNet with Spline fusion) was not as large as the gain achieved for the U-Net. This was expected: as the performance improves and gets closer to the upper limits of performance bounded by the aleatoric uncertainty, the levels of improvement achieved by individual components do not necessarily linearly add up. 

\begin{table}
\small
\centering
\caption{Performance metrics (on the MSSEG validation set) for different values of the hyperparameter $\beta$ used in training the 3D U-net on full-size images, and 3D patch-wise FC-DenseNet with different patch prediction fusion methods. The best values for each metric have been highlighted in bold. As expected, it is observed that higher $\beta$ led to higher sensitivity (recall) and lower specificity. The combined performance metrics, in particular APR, $F_2$ and DSC indicate that the best performance was achieved at $\beta=1.5$. Note that for highly unbalanced (skewed) data, the APR and $F_2$ score are preferred figures of algorithm performance compared to DSC ($F_1$ score), which is relatively insensitive and less representative of differences in unbalanced data. Therefore, while the difference in DSC was not significant for our most complex, best-performing methods (e.g. 3D patch-wise FC-DenseNet + Spline Fusion), the improvement in $F_2$ score was significant (70.5 vs. 71.6).}
\label{table:res}
 \begin{tabular}{|c||c|c|c|c|c|c|c|} 
 \hline
  & \multicolumn{5}{c|}{3D U-Net}\\
 \hline
$\beta$ & DSC & Sensitivity & Specificity & $F_2$ score & APR\\
 \hline
1.0 & 53.42 & 49.85 & \textbf{99.93} & 51.77 & 52.57 \\
 \hline
1.2 & 54.57 & 55.85 & 99.91 & 55.47 & 54.34 \\
 \hline
1.5 & \textbf{56.42} & 56.85 & 99.93 & \textbf{57.32} & \textbf{56.04} \\
 \hline
2.0 & 48.57 & 61.00 & 99.89 & 54.53 & 53.31 \\
 \hline
3.0 & 46.42 & \textbf{65.57} & 99.87 & 56.11  & 51.65 \\
 \hline
\end{tabular}

\begin{tabular}{|c||c|c|c|c|c|} 
 \hline
  & \multicolumn{5}{c|}{3D patch-wise FC-DenseNet + Tiling}\\
 \hline
$\beta$ & DSC & Sensitivity & Specificity & $F_2$ score & APR\\
 \hline
1.0 & 67.53 & 68.55 & \textbf{99.95} & 66.02 & 70.5 \\
 \hline
1.5 & \textbf{68.18} & 74.1 & 99.93 & \textbf{68.5} & \textbf{71.86} \\
 \hline
3.0 & 62.55 & \textbf{75.98} & 99.91 & 67.03  & 67.75 \\
 \hline
\end{tabular}

 \begin{tabular}{|c||c|c|c|c|c|} 
 \hline
  & \multicolumn{5}{c|}{3D patch-wise FC-DenseNet + Uniform Fusion}\\
 \hline
$\beta$ & DSC & Sensitivity & Specificity & $F_2$ score & APR\\
 \hline
1.0 & 68.81 & 75.28 & \textbf{99.94} & 69.91 & 72.15 \\
 \hline
1.5 & \textbf{68.99} & 79.97 & 99.90 & \textbf{71.96} & \textbf{73.08}\\
 \hline
3.0 & 63.05 & \textbf{83.55} & 99.89 & 70.65 & 69.85\\
 \hline
\end{tabular}

 \begin{tabular}{|c||c|c|c|c|c|} 
 \hline
  & \multicolumn{5}{c|}{3D patch-wise FC-DenseNet + Spline Fusion}\\
 \hline
$\beta$ & DSC & Sensitivity & Specificity & $F_2$ score & APR\\
 \hline
1.0 & \textbf{70.3} & 74.49 & \textbf{99.95} & 70.45 & 73.3 \\
 \hline
1.5 & 69.9 & 78.58 & 99.92 & \textbf{71.6} & \textbf{73.59}\\
 \hline
3.0 & 64.34 & \textbf{81.02} & 99.91 & 70.58 & 70.13\\
 \hline
\end{tabular}

\end{table}

\begin{figure}
    \centering
    \includegraphics[width=1\columnwidth]{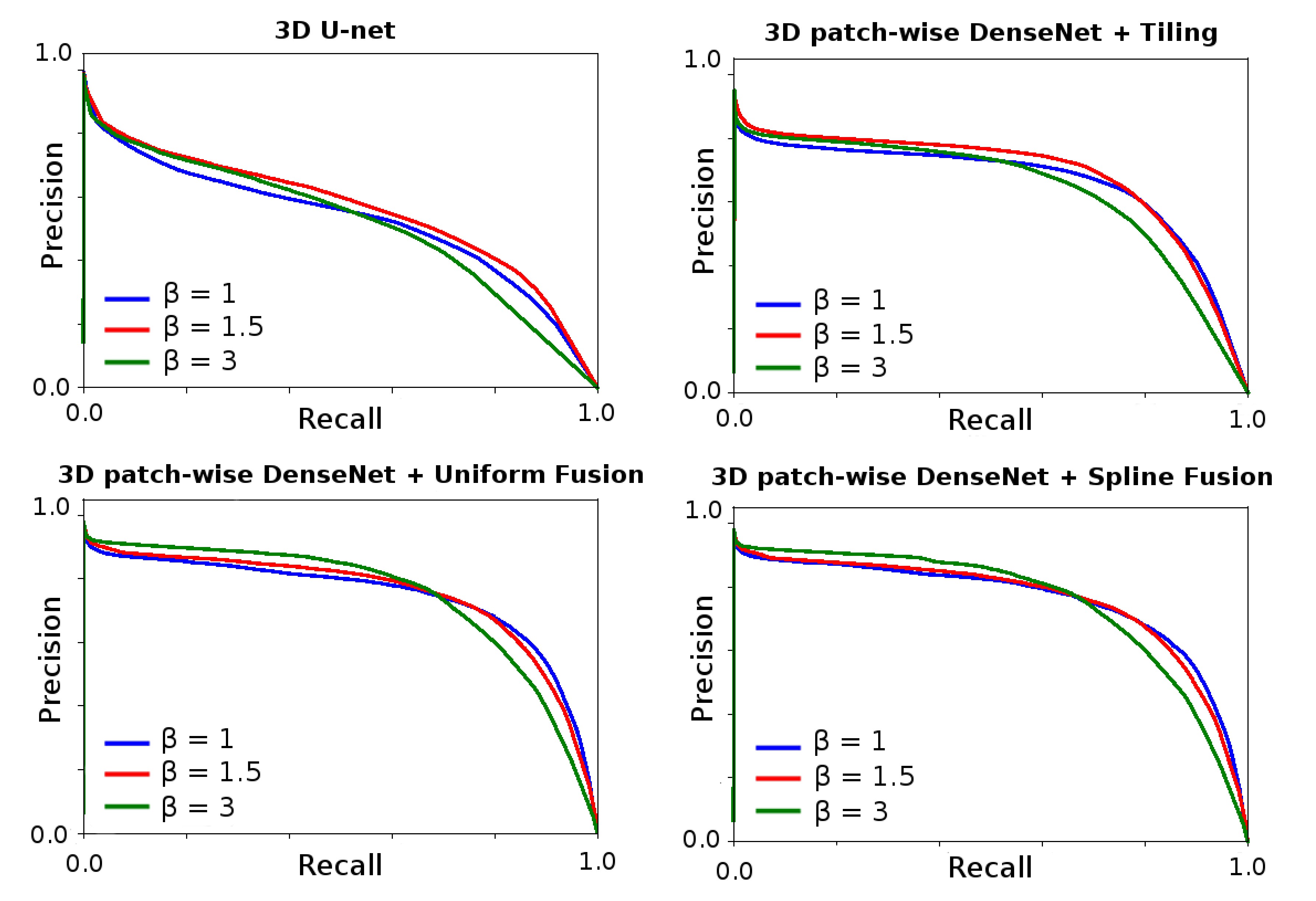}
    \caption{PR curves obtained by the four examined approaches with different loss function $\beta$ values on MSSEG. The best results based on the precision-recall trade-off were always obtained at $\beta=1.5$ and not with the Dice loss function ($\beta=1.0$); although the difference was less significant (bottom right plot) when we used large overlapping patches with our patch selection and patch prediction fusion methods that contributed to achieve better balanced sampling of data and improved fusion of augmented data. The combination of the asymmetric loss function and our 3D patch-wise FC-DenseNet with spline patch prediction fusion generated the best results (Table~\ref{table:res}). We emphasize that in case of unbalanced data, the PR curves are preferred performance criteria compared to ROC curves~\cite{boyd2013area,davis2006relationship,fawcett2006introduction}. Even though the differences may not seem large in these curves, all figures visualizing lesion segmentations in this article show that small differences in performance metrics (such as the Dice similarity coefficient) correspond to large visual differences between segmented lesions.}
    \label{fig:PR}
\end{figure}

\begin{figure}
    \centering
    \includegraphics[width=1\columnwidth]{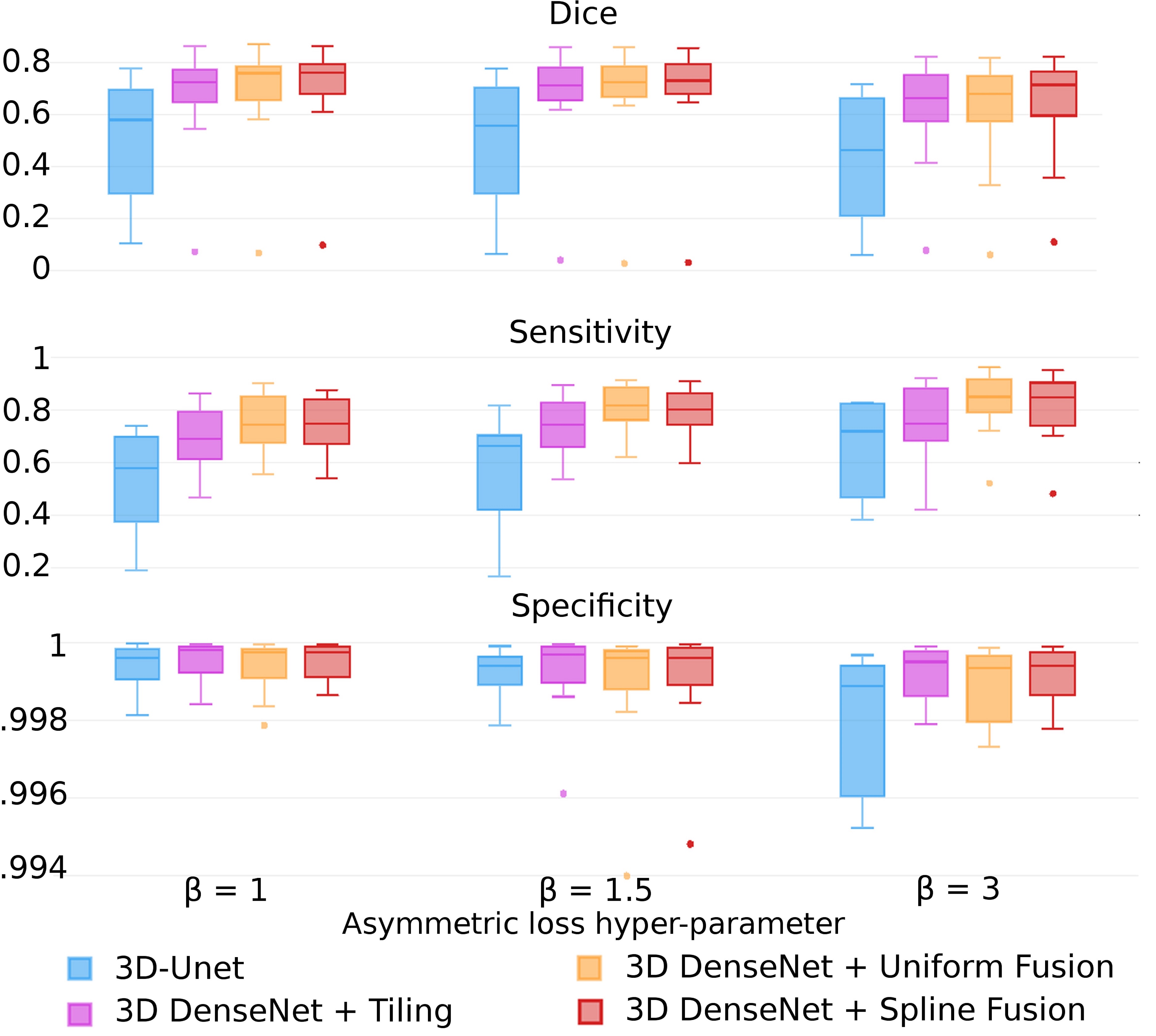}
    \caption{Boxplots of the MSSEG evaluation scores: Dice, sensitivity, and specificity for the four examined approaches. Overall, these results show that our FC-DenseNet model with the asymmetric loss function and spline patch prediction fusion made the best trade-off between sensitivity and specificity and generated the highest Dice coefficients among all methods.}
    \label{fig:Box}
\end{figure}

Figures~\ref{fig:HighL}, ~\ref{fig:MediumL} and~\ref{fig:LowL} show the effect of different hyper-parameter ($\beta$) values on segmenting a subject with high density of lesions, medium density of lesions and a subject with very few lesions, respectively. The improvement by using the asymmetric loss function was specifically significant in cases with very small number of lesion voxels as can be seen in Figure~\ref{fig:LowL}. Independent of the network architecture, training with the Dice loss function ($\beta=1$), resulted in a high number of false negatives as many lesions were missed. Note that a high value of $\beta=3$ also resulted in a drop in performance. Figure~\ref{fig:PR} shows the PR curves for three $\beta$ levels for the 3D U-Net and the 3D patch-wise FC-DenseNet with tiling, uniform fusion, and spline weighted fusion of patch predictions. As it can be seen in the PR curves (Figure~\ref{fig:PR}) and APR results in Table~\ref{table:res} for different architectures, the best results corresponding to a good trade-off between sensitivity (recall) and specificity was achieved using the asymmetric loss function with $\beta=1.5$. Figure~\ref{fig:Box} shows the boxplots of Dice, sensitivity, and specificity for the four networks trained with the loss function with different $\beta$ levels. Although, $\beta=1.5$ slightly decreased specificity, it led to a significant improvement in sensitivity (Figure~\ref{fig:Box}) and the APR, $F_1$ and $F_2$ scores (Table~\ref{table:res}). We further discuss the significance of these results in the MSSEG data in the Discussion section.

\subsubsection{Results on the ISBI challenge}
To further evaluate the performance of our proposed patch-wise 3D FC-DenseNet method with patch prediction fusion and the effect of the asymmetric loss function in making the trade-off between precision and recall, we compared our trained model with models trained with the Generalized Dice Loss (GDL) function~\cite{Sudre2017gdice} (using the default setting: squared reciprocal weight on the lesion class) and the focal loss function~\cite{lin2018focal} (using the reported best performing setting, $\alpha = 0.25$ and $\gamma = 2$) in MS lesion segmentation. For the asymmetric similarity loss function, we trained our model with our best performing $\beta$ value of 1.5 (based on MSSEG evaluation) on the ISBI dataset.

The challenge test results of our 3D patch-wise FC-DenseNet trained with the asymmetric loss function, the focal loss, and the GDL, with the patch selection and spline-weighted patch prediction fusion on the ISBI challenge is shown in Table~\ref{table:challenge}. As demonstrated in the table, according to the challenge overall score, as of September 2018, we ranked $1^{st}$ with our model trained with the focal loss, $6^{th}$ with our model trained with the GDL, and $8^{th}$ with our model trained with the $F_\beta$ loss function. The overall score, however, is not necessarily an optimal weighted average of all evaluation scores; therefore to gain insight into the characteristics and performance of each model trained purposefully with the loss functions, one should look into specific evaluation metrics.

\begin{table*}[ht!]
\small
\centering
\caption{The top ten ranking teams of the ISBI longitudinal MS lesion segmentation challenge (\textcolor{blue}{https://smart-stats-tools.org/lesion-challenge}) as of September 2018 with average metrics of challenge score, Dice coefficient, Jaccard coefficient, positive predictive value (PPV), sensitivity (TPR), lesion TPR based on lesion count (LTPR), lesion FPR based on lesion count (LFPR), Volume Difference (VD), Average Symmetric Surface Difference (SD) and average segmentation volume. Average manual volume of the two raters in the challenge was \textbf{15,648}. Our proposed method (IMAGINE) with $F_\beta$ achieved the best results in 4 out of 9 evaluation metrics, and second highest in DSC and Jaccard compared to the other methods, while our patch-wise 3D FC-DenseNet trained with focal loss achieved the first place as well as better results in PPV and LFPR metrics. For some of the most recent submissions, UVA, Unige, PAVIS and Braz, we could not find any published articles for reference. In case of multiple submissions by same group, we reported their highest ranked results. It is noteworthy that the performance of methods with the overall challenge score of over 90 is considered to be comparable to human raters.}
\label{table:challenge}
 \begin{tabular}{|c||c|c|c|c|c|c|c|c|c|c|c|} 
 \hline
 & Score & DSC & Jaccard & PPV & TPR & LTPR & LFPR & VD & SD & Avg Segm Vol & Date\\
 \hline
IMAGINE (Focal) & \textbf{92.486} & 58.41 & 43.08 & \textbf{92.07} & 45.58 & 41.35 & \textbf{8.66} & 49.72 & 4.48 & 8176 & 06/2018\\
\hline
UVA (best) & 92.402 & \textbf{67.37} & \textbf{52.03} & 83.22 & 60.03 & 48.043 & 17.24 & 34.79 & 3.07 & 11390 & 08/2018 \\
\hline
Unige (best) & 92.118 & 61.13 & 45.65 & 89.91 & 49.00 & 41.03 & 13.93 & 45.37 & 3.68 & 8768 & 08/2018 \\
\hline
PAVIS (best) & 92.108 & 62.75 & 47.25 & 88.41 & 51.38 & 41.67 & 14.69 & 42.17 & 3.42 & 9233 & 09/2018\\
 \hline
asmsl~\cite{andermatt2017lesion} & 92.076 & 62.98 & 47.38 & 84.46 & 53.69 & 48.7 & 20.13 & 40.45 & 3.65 & 10532 & 02/2017\\
\hline
IMAGINE (GDL) & 91.817 & 61.04 & 45.16 & 86.01 & 49.66 & 35.03 & 9.46 & 42.07 & 3.84 & 8558 & 08/2018\\
\hline
Braz (best) & 91.735 & 60.28 & 44.66 & 90.01 & 47.75 & 33.53 & 10.81 & 46.45 & 3.97 & 8353 & 05/2018\\
 \hline
IMAGINE ($F_\beta$) & 91.523 & 65.74 & 50.04 & 71.39 & \textbf{66.77} & \textbf{50.88} & 21.93 & 37.27 & \textbf{2.88} & \textbf{14429} & 04/2018\\
 \hline
nic vicorob test & 91.440 & 64.28 & 48.52 & 79.24 & 57.02 & 38.72 & 15.46 & \textbf{32.58} & 3.44 & 10269 & 02/2017\\
 \hline
VIC TF FULL & 91.331 & 63.04 & 47.21 & 78.66 & 55.46 & 36.69 & 15.29 & 33.84 & 3.56 & 10740 & 05/2017\\
 \hline
MIPLAB v3 & 91.267 & 62.73 & 47.13 & 79.96 & 54.98 & 45.39 & 23.17 & 35.85 & 2.91 & 10181 & 08/2016\\
 \hline
\end{tabular}

\end{table*}

Our model trained with the focal loss achieved the best PPV and LFPR scores among all submitted methods indicating its excellent performance in minimizing lesion false positive rate in a trade-off to keep other performance metrics at desirable levels. On the other hand, our network trained with the asymmetric similarity loss function achieved better results than the top ranking teams in 4 out of 9 evaluation metrics, namely TPR, LTPR, SD, and the average segmentation volume. It is noteworthy that while purposefully shifting the emphasis towards higher true positive rates, this model achieved competitive overall DSC and Jaccard indices and the lowest surface distance. We note that while our main goal was to achieve high recall (sensitivity - TPR) in using the asymmetric loss function, which was accomplished, we also achieved the best estimation of average lesion volume and the closest lesion surface distance among all submitted methods. This was not unexpected and showed that the data imbalance was effectively addressed and the trained network performed well on the test set. The significance of these results are further discussed in the Discussion section.

Figure~\ref{fig:ISBI} shows the true positive, false negative, and false positive voxels overlaid on axial views of the baseline scans of two patients with high and low lesion loads (top and bottom two rows, respectively) from our cross-validation folds in the ISBI challenge experiments. These results show low rate of false negatives in challenging cases.
Lesion volume sizes expanded from 3.5K to 38K in the ISBI training and validation sets, while the challenge test set had a range between 1K to 53K number of lesion voxels per 3D image. This shows the wider range of lesion cases in the challenge; testing the generalizablility attribute of the submitted methods. Subjects 2 and 3 in the validation set shown in Figure \ref{fig:ISBI}, have lesion voxel sizes of 32K and 11K, respectively.

\section{Discussion}
\label{sec_discussion}

Experimental results in MS lesion segmentation (Table~\ref{table:res}) show that almost all performance evaluation metrics (on test data) improved by using an asymmetric similarity loss function rather than using the Dice similarity in the loss layer. While the loss function was deliberately designed to weigh recall higher than precision (at $\beta=1.5$), consistent improvements in all test performance metrics including DSC and $F_2$ scores on the test set indicate improved generalization through this type of training. Compared to DSC which weighs recall and precision equally, and the ROC analysis, we consider the area under the PR curves (APR, shown in Figure~\ref{fig:PR}) the most reliable performance metric for such highly skewed data~\cite{fawcett2006introduction,boyd2013area}.

\begin{figure*}
    \centering
    \includegraphics[width=0.97\textwidth]{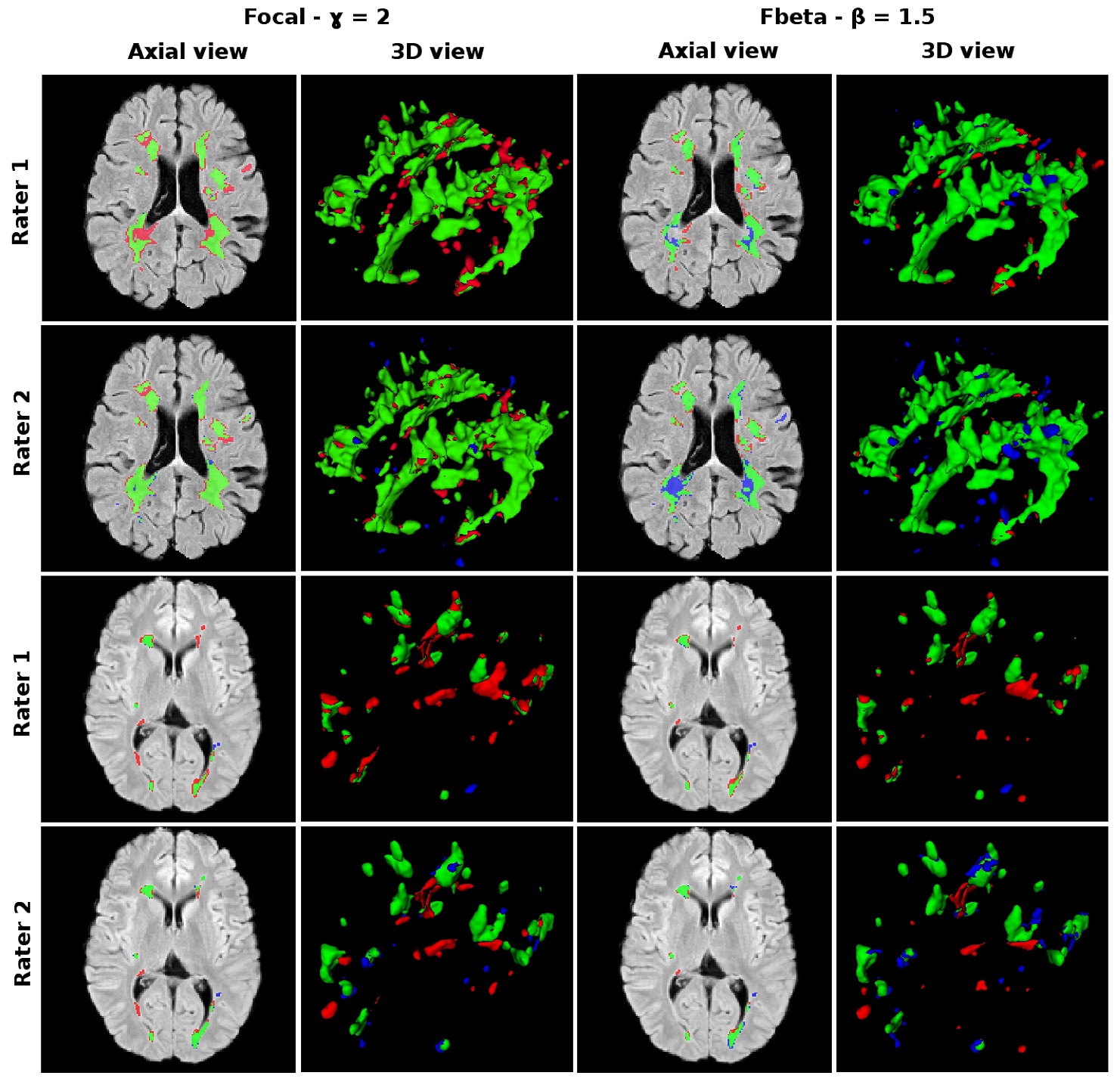}
    \caption{ISBI validation results for focal and $F_\beta$ loss functions. High volume of lesion (top two rows) and Low volume of lesion (bottom two rows) segmentation results compared to both manual segmentations (GTs) for baseline scans of patient 2 and patient 3, respectively. Computed $F_\beta$ DSC scores of 82.35, 79.62, 71.9 and 74.47 was calculated from top to bottom respectively. Computed focal DSC scores of 81.1, 82.66, 64.54, 77.48 was calculated from top to bottom respectively. True positives, false negatives and false positives are colored in the order of green, blue and red. The results show that the focal loss model predicts less false negatives with the expense of more false positives in general. We chose these subjects based on the roughly equal average DSC scores of 77.09 vs 76.45 for $F_\beta$ and focal approaches, respectively.}
    \label{fig:ISBI}
\end{figure*}

Table~\ref{table:res} shows relative improvements due to the use of different network architectures, loss functions, and fusion strategies, based on our proposed algorithm, that collectively resulted in the best lesion segmentation performance. It also shows that those relative improvements do not linearly add up, since as the performance improves, we get closer to the upper limits of performance bounded by aleatoric uncertainty. We should also note that in such unbalanced problems, even small differences in performance metrics do project into rather large and important differences in results, especially in medical applications, as is observed in the examples shown in Figure~\ref{fig:ISBI} for the focal loss and $F_\beta$ results; where the difference between average DSC score of the two methods for this subject was about 0.6\% (77.09 vs. 76.45) despite the huge differences in lesion detection performance. In fact, as we mentioned throughout the article and in the results section (Table~\ref{table:res}), for such unbalanced data, $F_2$ score is a much better performance metric compared to the DSC ($F_1$ score) as it makes a better balance between precision and recall.

With our proposed 3D patch-wise FC-DenseNet method we achieved improved precision-recall trade-off and high average DSC scores of 69.9\% and 65.74\% which are better than the highest ranked techniques examined on the MSSEG 2016 and ISBI challenges, respectively. In the MSSEG challenge the 1st ranked team~\cite{mckinley2016lesion} reported an average DSC of 67\%, and the 4th ranked team~\cite{vera2016lesion} reported an average DSC of 66.6\%. In the ISBI challenge we ranked higher than the top ten teams in 4 out of 9 evaluation metrics, and second highest in DSC and Jaccard metrics, with our proposed asymmetric similarity loss function namely $F_\beta$, and ranked $1^{st}$ (based on the overall score) with our model trained with the focal loss as we achieved the best PPV and LFPR metrics among all examined methods (Table~\ref{table:challenge}). We achieved improved performance by using a 3D patch-wise FC-DenseNet architecture together with asymmetric loss functions and our patch prediction fusion method.

For consistency in comparing to the literature on these challenges we reported all performance metrics, in particular DSC, sensitivity, and specificity for MSSEG, and nine metrics as well as the overall score for ISBI. We were able to balance between different performance metrics with various loss functions. When a model is trained to be used as a clinical decision support system, where detected lesions are reviewed and confirmed by experts, recall (TPR), the $F_2$ scores, and in particular the LTPR are more important figures than PPV, the $F_1$ score, and the LFPR. Expert manual segmentation of the full extent of lesions (used as ground truth) is very challenging. The detection and count of small lesions, on the other hand, is paramount in MS diagnosis and prognosis as new lesion count as well as lesion classification (to active and inactive) is used in Disease-Modifying Treatment (DMT)~\cite{giovannoni2018disease} algorithms. Measures based on lesion count, such as LTPR and LFPR, are thus considered more important metrics than TPR, PPV, and DSC. We achieved the highest LTPR and the lowest LFPR among other methods in the ISBI challenge with our 3D patch-wise FC-DenseNet trained with our proposed asymmetric loss function and the focal loss, respectively, while achieving top performance according to the combination of all metrics. Clinical judgment remains an integral part of current DMT algorithms. To guide a decision support system we recommend our model trained with the proposed asymmetric similarity loss function as it generated the best LTPR value (Table~\ref{table:challenge}) among all other methods. 

\section{Conclusion}
\label{sec_conclusion}

To effectively train deep neural networks for highly unbalanced lesion segmentation in medical imaging, we added asymmetric loss layer to two state-of-the-art 3D fully convolutional deep neural networks based on the DenseNet~\cite{huang2017dense,jegou2017tiramisu} and U-net (V-net) architectures~\cite{ronneberger2015u,milletari2016v}. To work with any-size 3D input images and achieve intrinsic data augmentation and balanced sampling to train our FC-DenseNet architecture with similarity loss functions, we proposed a patch selection and augmentation strategy, and a patch prediction fusion method based on spline-weighted soft voting. We achieved marked improvements in several important evaluation metrics by our proposed method in two competitive challenges, outperforming state-of-the-art methods, and achieving top performance in MS lesion segmentation through September 2018. We compared performance of networks trained with loss functions based on Dice similarity coefficient, generalized Dice (GDL)~\cite{Sudre2017gdice}, focal loss~\cite{lin2018focal}, and asymmetric loss function based on $F_\beta$ scores according to several performance metrics. To put the work in context, we reported average DSC, $F_2$, and APR scores of 69.9, 71.6, and 73.59 for the MSSEG challenge, and average DSC, Jaccard and Sensitivity (TPR) scores of 65.74, 50.04 and 66.77 for the ISBI challenge respectively, which, along with the publicly available test results on ISBI challenge website, indicate that our approach performed better than the latest methods applied in MS lesion segmentation~\cite{commowick2016msseg,carass2017longitudinal,andermatt2017lesion,valverde2017improving,mckinley2016lesion,vera2016lesion}. Based on these results, we recommend the use of asymmetric similarity loss functions within our proposed method based on large overlapping image patches and patch prediction fusion to achieve better precision-recall balancing in highly unbalanced medical image segmentation applications.

\ifCLASSOPTIONcaptionsoff
  \newpage
\fi

\bibliographystyle{IEEEtran}
\bibliography{main}

\EOD
\end{document}